\documentclass[final,12pt]{elsarticle}
\usepackage{amsthm}
\usepackage{longtable}
\usepackage{hyperref}
\usepackage{setspace}
\usepackage{amsmath, soul}

\usepackage{amssymb}

\usepackage{lineno,hyperref}	
\usepackage{xcolor,soul, booktabs}
\usepackage{caption}
\usepackage{subcaption}
\newcommand{\specialcell}[2][t]{%
  \begin{tabular}[#1]{@{}l@{}}#2\end{tabular}}

\usepackage{epstopdf}

\journal{Journal of Food Engineering}
\begin{document}

\begin{frontmatter}

\title{Shedding Light on the Ageing of Extra Virgin Olive Oil: Probing the Impact of Temperature with Fluorescence Spectroscopy and Machine Learning Techniques}



\author[zhaw,toelt]{Francesca Venturini\corref{mycorrespondingauthor}}
\ead{vent@zhaw.ch}

\author[zhaw]{Silvan Fluri}
\author[idsia]{Manas Mejari}
\author[zhaw]{Michael Baumgartner}
\author[idsia]{Dario Piga}
\author[toelt,hslu]{Umberto Michelucci}

\cortext[mycorrespondingauthor]{Corresponding author}

\address[zhaw]{Institute of Applied Mathematics and Physics, Zurich University of Applied Sciences, Technikumstrasse 9, 8401 Winterthur, Switzerland}
\address[toelt]{TOELT LLC, Advanced AI Lab, Birchlenstrasse 25, 8600 D\"ubendorf, Switzerland}
\address[idsia]{IDSIA - Dalle Molle Institute for Artificial Intelligence, USI-SUPSI, Via la Santa 1, 6962 Lugano, Switzerland}
\address[hslu]{Lucerne  University of Applied Sciences and Arts, Computer Science Department, Lucerne, Switzerland}

\begin{abstract}

This work systematically investigates the oxidation of extra virgin olive oil (EVOO) under accelerated storage conditions with UV absorption and total fluorescence spectroscopy. With the large amount of data collected, it proposes a method to monitor the oil's quality based on machine learning applied to highly-aggregated data.
EVOO is a high-quality vegetable oil that has earned worldwide reputation for its numerous health benefits and excellent taste. Despite its outstanding quality, EVOO degrades over time owing to oxidation, which can affect both its health qualities and flavour.  Therefore, it is highly relevant to quantify the effects of oxidation on EVOO and develop methods to assess it that can be easily implemented under field conditions, rather than in specialized laboratories.

The following study demonstrates that fluorescence spectroscopy has the capability to monitor the effect of oxidation and assess the quality of EVOO, even when the data are highly aggregated. It shows that complex laboratory equipment is not necessary to exploit fluorescence spectroscopy using the proposed method and that cost-effective solutions, which can be used in-field by non-scientists, could provide an easily-accessible assessment of the quality of EVOO.

\end{abstract}


\end{frontmatter}


\section{Introduction}
\label{sec:intro}

Extra virgin olive oil (EVOO) is a high-quality product that is widely consumed due to its health benefits and culinary properties. Its historically predominant Mediterranean diffusion is spread to non-producing countries, and EVOO is becoming more popular globally. The assurance of the quality of EVOO, from the production throughout its shelf life,  is therefore becoming of great importance. The quality of EVOO depends on multiple factors, such as the olive cultivar, the geographic location, the processing method, and storage conditions, just to name the most important ones.

During storage, EVOO undergoes chemical and physical changes that can result in the deterioration of its quality attributes, such as flavour, aroma, and colour, but also in the loss of nutritional properties. The main causes of this degradation are autoxidation (in darkness) and photo-oxidation (in presence of light) which depend on temperature, time, amount of available oxygen or water, and result in the degradation of fatty acids, tocopherols and chlorophylls and in formation of primary and secondary oxidation products\cite{gomez2004evolution,mancebo2008kinetic,aparicio2010thermal,esposto2017effect,lam2020thermal}.

The quantitative assessment of the effects of ageing during storage and the monitoring of the quality of EVOO throughout its entire life cycle is crucial to ensure the high quality of the oil is guaranteed until the expiration time. Previous studies indicate that the concentration of several compounds change during storage and that the freshness is significantly influenced by the storage conditions and not only by the time elapsed from the harvest of the olives
\cite{fadda2012changes,aparicio2017predicting,conte2020temperature}. The compounds studied include chlorophylls, carotenoids, tocopherols and lipids \cite{aparicio2012thermal,camerlingo2019surface,botosoa2021front,lopes2023evaluation}. 

According to the the European Regulation and its amendments \cite{regulation1991commission,european2013commission} the quality assessment of EVOO is performed through a series of chemical analysis and on sensory evaluation by certified laboratories and  panelists respectively. These analyses are time-consuming, expensive, and are not easily accessible for a large number of producers.
Alternatively to chemical analysis, this work investigates the use of fluorescence spectroscopy, which has the advantages of being a rapid, cost-efficient, and at the same time sensitive technique \cite{el2020rapid,guzman2015evaluation}. Fluorescence spectroscopy has been demonstrated to be a technique that can be used successfully to monitor the freshness of EVOO and the quality of olive oil in general  because of the natural presence of fluorophores in olive oil, the strongest of which are chlorophyll \cite{sikorska2008fluorescence,karoui2011fluorescence,lobo2020monitoring,al2021cultivar}. Machine learning (ML) has emerged as a promising tool for the extraction from spectral analysis of relevant features and for the objective and efficient evaluation of the quality of EVOO \cite{ordukaya2017quality,vega2020deep,venturini2021exploration,zaroual2021application}. This is because ML algorithms can learn from data to identify patterns and classify samples based on their quality attributes.

This study aims to contribute to a general understanding of EVOO ageing by systematically studying the effects of thermal degradation intended as accelerating ageing condition. This is done by a comprehensive spectroscopic analysis of the absorption and fluorescence properties. To observe general trends in ageing, a high sample variability was chosen in terms of geographical origin and composition properties. The quality assessment of EVOO is performed by measuring the UV-absorption parameters $K_{268}$ and $K_{232}$ defined by the European Regulations \cite{regulation1991commission,european2013commission}. It should be noted that although the EU defined quality assessment includes other parameters not measured by this study, if an oil exceeds the limits in terms of UV parameters, it can definitely not be sold as EVOO any more.

The use of ML methods is being explored to determine whether simple fluorescence sensors can be used to qualitatively control oil quality in the field and at production and storage sites. Therefore, the large amount of data is analyzed in an aggregated form. An analysis of the spectral information is expected to provide additional information. However, the theses put forward in this work is that a sensor measuring the fluorescence intensity at one or maximum two excitation wavelengths can reliably and robustly predict if the EVOO is still in this quality class. The advantage of such a sensor would be that, requiring only one or two LEDs for the excitation and a photodiode for the detection, its design, dimensions and working principle would be extremely simple and cheap. Such a sensor, which could be the subject of future work, would open the way to a fast and easy qualitative assessment of the quality of olive oil.

The contributions of this work are four: 1) extensive UV-absorption spectroscopy measurement at the wavelength specified by the the European Regulations over time for commercial EVOO; 2) extensive fluorescence emission analysis of the same oils through the acquisition of excitation-emission matrices (EMMs); 3) identification of the two wavelengths for which the overall relative change in the fluorescence is maximal; 4) proposal of a method for the prediction of olive oil quality based on aggregated data.

\section{Materials and Methods}
\label{sec:mame}

\subsection{Extra Virgin Olive Oil Samples}
\label{sec:samples}

This study involves 24 different commercially available extra virgin olive oils. The oils were chosen to cover a wide range of price classes and production regions (Italy, Spain, Portugal, Greece, and unspecified in Europe). The list of the EVOOs is reported in Table \ref{tab:oils}. 

\begin{table}[t!]
\begin{tabular}{c|l|l}
\specialrule{.2em}{0em}{0em} 
\textbf{Label} & \textbf{Sample description} & \textbf{Origin}\\
\specialrule{.1em}{0em}{0em}
A & \specialcell{Coop Naturaplan Italienisches Oliven\"ol (BIO)} & IT\\
B & Hacuinda Don Paolo & ES\\
C & Monocultivar Nocellara Bio & IT\\
D & Monini, Toscano IGP & IT, Tuscany\\
E & Monini, Classico & IT\\
F & Oliva, Favola & IT\\
G & Migros, M Classics & ES\\
H & Alexis, Manaki & GR\\
I & Migros, Bio Italienische Olivenöl & IT\\
J & Alnatura, Natives Olivenöl extra & ES\\
K & Migros, Bio Griechisches Olivenöl & GR\\
L & Demeter, Spanisches Olivenöl & ES\\
M & Filippo Berio, Il Classico & EU\\
N & \specialcell{Demeter, Bio Coop Naturaplan\\Portugisisches Olivenöl} & PT\\
O & Castillo, Don Felpe & ES\\
P & Coop, Naturaplan Bio Griechisches Olivenöl & GR\\
Q & Demeter, Son Naava & ES, Mallorca\\
R & Iliada, Kalamata PDO & GR\\
S & Sapori d'Italia, Sicilia & IT, Sicily\\
T & San Giuliano, Sardegna DOP & IT, Sardinia\\
U & Coop, Naturaplan Bio Spanisches Olivenöl & ES\\
V & San Giuliano, Fruttato & IT\\
W & San Giuliano, L'Originale & IT\\
X & Coop, Qualité-Prix & IT, ES, GR\\
\specialrule{.1em}{0em}{0em}
\end{tabular}
\caption{List of the olive oils samples analyzed in this study and their geographycal origin. IT: Italy, ES: Spain, GR: Greece, PT: Portugal, EU: European not specified.
\label{tab:oils}}
\end{table}

The oxidative stability of olive oils was investigated under accelerated storage conditions, following the modified Schaal oven test Celsius \cite{evans1973} by holding the samples at 60$^{\circ}$ degree in the dark. This approach enables to analyze the oxidation in shorter time because it reproduces oxidative changes similar to those observed under actual shelf life conditions \cite{apariciobook}.

All oils were measured fresh and at nine different time intervals (ageing stages) up to a maximum of 53 days in the oven. The exact duration of the aging for each ageing step is reported in Table \ref{tab:ageing}.

\begin{table}[t!]
  \centering
  \begin{tabular}{c|c}
  \specialrule{.2em}{0em}{0em} 
    \textbf{Ageing step} & \specialcell{\textbf{Ageing Duration}\\\textbf{at 60$^\circ$ C (days)} } \\ 
    \specialrule{.1em}{0em}{0em}
    0 & fresh \\
    1 & 2 \\
    2 & 4 \\
    3 & 7 \\
    4 & 9 \\
    5 & 18 \\
    6 & 27 \\
    7 & 36 \\
    8 & 45 \\
    9 & 53 \\
  \specialrule{.1em}{0em}{0em}
  \end{tabular}
  \caption{Ageing duration in days for each ageing step.\label{tab:ageing}}
\end{table}

The samples were prepared from commercial bottles by filling to the top and hermetically closing 4 ml vials to minimize the amount of oxygen in the headspace.
One vial per measurement was prepared: three samples for each of the 24 oils and for 10 ageing steps for a total of 720 samples. This procedure allowed for identical ageing conditions for all oils and stages investigated. Otherwise progressively larger amount of oxygen would result from removing part of the oil for the measurement.
In facts, it was shown that oxygen in the headspace influences the oxidation and results in an increase in the concentration of free fatty acids and peroxide value \cite{iqdiam2020influence}. At each ageing step, three identical samples of the 24 oils, for a total of 72 samples, were taken from the oven and measured. The redundancy was done to have multiple measurements of the same oil at any ageing conditions to check the reproducibility of the results.
More details on EVOOs, sample preparation, and the ageing process are reported in the article  
\cite{venturini2023}.

\subsection{UV-Absorption and Fluorescence Spectroscopy Measurements}
\label{sec:measurement}

The quality of EVOOs at the various phases of ageing can be monitored using UV absorption spectroscopy. The European Regulation and its amendments \cite{regulation1991commission,european2013commission} define precise limits for three parameters (called extinction coefficients) that quantify the absorbance in the UV, where the oxidation products absorb the most. These parameters are $K_{232}$, the absorbance at 232 nm, $K_{268}$, the absorbance at 268 nm, and $\Delta K$ determined from the absorbance at 264 nm, 268 nm and 272 nm according to the formula
\begin{equation}
    \Delta K = \left[
        K_{268}-\left(
            \frac{K_{264}-K_{272}}{2}
        \right)
    \right]
\end{equation}

The UV spectroscopy extinction coefficients were determined with an Agilent Cary 300 UV-Vis spectrophotometer on diluted samples. Both the sample preparation and the measurement method followed the procedures defined by the European Regulation and its amendments \cite{regulation1991commission,european2013commission}. The three parameters $K_{232}$, $K_{268}$, and $\Delta K$ were measured for each type of oil and ageing step at a constant temperature of 22 $^{\circ}$C. 

The total fluorescence spectroscopy characteristics of the 720 samples were determined by acquiring the excitation emission matrix (EEM). The EEMs were measured with an Agilent Cary Eclipse Fluorescence Spectrometer by changing the wavelength of the illuminating excitation light from 300 to 650 nm in steps of 10 nm and measuring the intensity of the fluorescence emitted as a function of the wavelength between 300 and 800 nm in steps of 2 nm. All measurements were made at a constant temperature of 22 $^{\circ}$C on undiluted samples. The detailed information on the acquisition parameters and procedures of both spectroscopic techniques are reported in \cite{venturini2023}. The complete dataset including both UV-absorption and fluorescence spectroscopy raw measurements is publicly available \cite{venturinidatasetares}.

\subsection{Machine Learning Methods}
\label{sec:mlm}

This work aims to determine whether it is possible to predict if an oil aged, for example after a long time from processing or in the bottle, is still extra virgin using only aggregated information from fluorescence spectra. The goal is to explore whether a very simple sensor based on the measurement of the fluorescence intensity, thus with a non-destructive purely optical method on undiluted samples, could be used to perform a quality assessment. The motivation for the aggregation of the data is to explore the potentiality of a low-cost and portable device with a hardware as simple as possible: one or at most two LEDs for the excitation and a single photodiode for the fluorescence intensity detection. Clearly the aggregation of the information of such a device poses a serious challenge as compared to the analysis of the entire EMMs.

Therefore, the proposed approach first requires the identification of the excitation wavelengths containing the largest amount of information for this specific purpose (Section \ref{sec:wavelength}). Once the method determines the most relevant wavelengths, the machine learning algorithm classifies the oil as EVOO or non-EVOO according to the UV-spectroscopy criteria of the European regulation \cite{regulation1991commission,european2013commission} (Sections \ref{sec:model} and \ref{sec:validation}).

\subsubsection{Information Content Maximization Approach}
\label{sec:wavelength}

To identify the fluorescence spectra that contain the most information about thermal degradation consists in determining at which excitation wavelength $\lambda_i$ $i=1,...,35$ the resulting fluorescence spectrum shows the greatest change between different ageing steps. The change is quantified mathematically with the Relative Error ($RE$) for a sample $j$, an excitation wavelength $\lambda_i$ and an ageing step $k$ that is defined as
\begin{equation} \label{eqn:RE}
    RE_{j,k}(\lambda_i) = \frac{\left\| I_{j,0}(\lambda_i) - I_{j,k}(\lambda_i) \right\|_2^2}{\left\| I_{j,0}(\lambda_i) \right\|_2^2},
\end{equation}
where $I_{j,k}(\lambda_i)$ is a vector whose components represent the intensity at various emission wavelengths resulting from excitation at the wavelength $\lambda_i$, for sample $j$ and aging step $k$.
$\left\| \cdot \right\|_2^2$ indicates the Euclidean norm squared, or in other words, the squared sum of all the values of the vector. 11 pixels of the measured spectra around the excitation frequency were set to zero (5 on the left, 5 on the right of the excitation frequency and the excitation frequency itself)  to remove the intensity due to Rayleigh scattering.

To determine the two wavelengths $\lambda^{[1]}$ and $\lambda^{[2]}$ for which the $RE$ for $k=9$ is largest and second larges, respectively, the following steps were taken. First the wavelength $\lambda^{[1]}_{j}$ for which the $RE$ for $k=9$ is largest was determined for each oil $j$. Then, the number of times that this wavelength appear in the list was counted. Finally, the two wavelengths that are most common are selected. Since the fluorophores in olive oil have broad absorption bands, wavelengths with variations of +/- 10 nm excite the same fluorophores. Therefore, the wavelength 310 nm can be counted as ``equivalen'' to 300 nm.

\subsubsection{Classification model}
\label{sec:model}

To classify the sample in EVOO or non-EVOO the following algorithms were tested in this work:
\begin{itemize}
    \item AdaBoost (with a decision tree as base model initialized with maximum depth of 1, with 5 estimators and a different random seed each training run. The algorithm was trained with a learning rate of 1 and the SAMME.R boosting algorithm \cite{hastie2009multi}).
    \item Random forest (with 100 estimators and maximum depth of 2).
    \item Logistic regression.
    \item Na\"ive Bayes.
\end{itemize}

The results reported in this paper are for the AdaBoost algorithm, as this proved to be the one with the best results for both parameters $K_{268}$ and $K_{232}$. The results obtained using the other algorithms are reported in full detail in the additional material.

The algorithms classify the oils following the criteria for the UV-spectroscopy limits for either the parameter $K_{268}$ or the parameter $K_{232}$ for EVOO defined by the European regulation:
\begin{itemize}
    \item class 1 if $K_{268}<0.22$ or class 0 if $K_{268}>0.22$.
    \item class 1 if $K_{232}<2.5$ or class 0 if $K_{232}>2.5$.
\end{itemize}
The parameter $\Delta K$ was not used for the classification since the error on the measurements is comparable to the overall change observed (see Section \ref{sec:results_UVAbs}). 
The dataset is unbalanced since after 53 days of ageing only seven oils are bad  using the $K_{232}$ criterion and only four oils are still good using the $K_{268}$ criterion (see Section \ref{sec:results_UVAbs}). Therefore, a random oversampling without replacement was used for the model training \cite{hastie2009multi}.

\subsubsection{Validation}
\label{sec:validation}

To validate the model, two different strategies were followed. These are summarised in Table \ref{tab:models}. The approaches reflect the classical validation methods used in time series \cite{fan2003nonlinear, mcquarrie1998regression}, since the  measurements were made at subsequent time points.

\begin{table}[!h]
    \centering
    \begin{tabular}{c||c||c }
         \specialrule{.2em}{0em}{0em} 
\begin{tabular}{@{}c@{}}Input ageing steps \\ used for training\end{tabular} 
 & \begin{tabular}{@{}c@{}}Validated on ageing \\ step (method 1)\end{tabular}   & \begin{tabular}{@{}c@{}}Validated on ageing \\ step (method 2)\end{tabular}  \\
\specialrule{.1em}{0em}{0em}
         4, 5  & 9 & 6 \\
         4, 5, 6  & 9 & 7  \\
         4, 5, 6, 7  & 9 & 8 \\
         4, 5, 6, 7, 8  & 9 & 9  \\
    \end{tabular}
    \caption{An overview of the different training methods used. For both methods an increasing number of ageing steps for training was used. In Method 1 the validation is done on step 9; in Method 2 the validation is done on the immediate next ageing step.}
    \label{tab:models}
\end{table}


For both methods, an increasing number of ageing steps was used for the training to determine the the amount if information needed for a reliable classification. In Method 1 the validation (prediction of the class) is always done on the 9$^{\textrm{th}}$ ageing step, and in Method 2 the class at the immediate next ageing step is predicted. For example, in Method 1 the $RE$s calculated from the spectra at ageing steps 4 and 5 was used for training, and the model was validated using the $RE$s corresponding to the ageing step 9, while in Method 2 the the model was validated using the $RE$s at ageing step 6.

\section{Results and Discussion}
\label{sec:results}

\subsection{UV-Absorption Results}
\label{sec:results_UVAbs}

\begin{figure}[hbt]
    \centering
    \includegraphics[width=\textwidth]{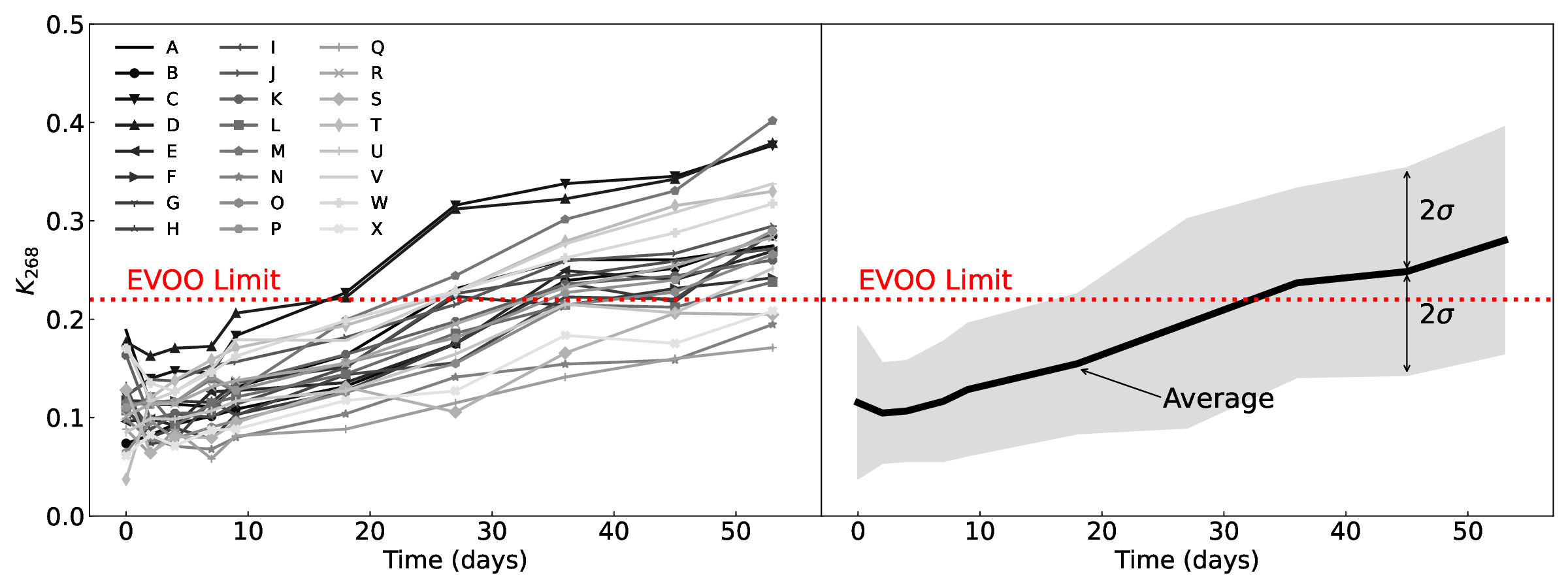}
    \caption{Left panel: UV-Spectroscopy $K_{268}$ parameter over the time during the ageing. Right panel: average of the values of all the oils and variation during the ageing. $\sigma$ is the standard deviation at each ageing step. The red line is the limit of $K_{268}$ according to the European Regulation.}
    \label{fig:K268}
\end{figure}

\begin{figure}[hbt]
    \centering
    \includegraphics[width=\textwidth]{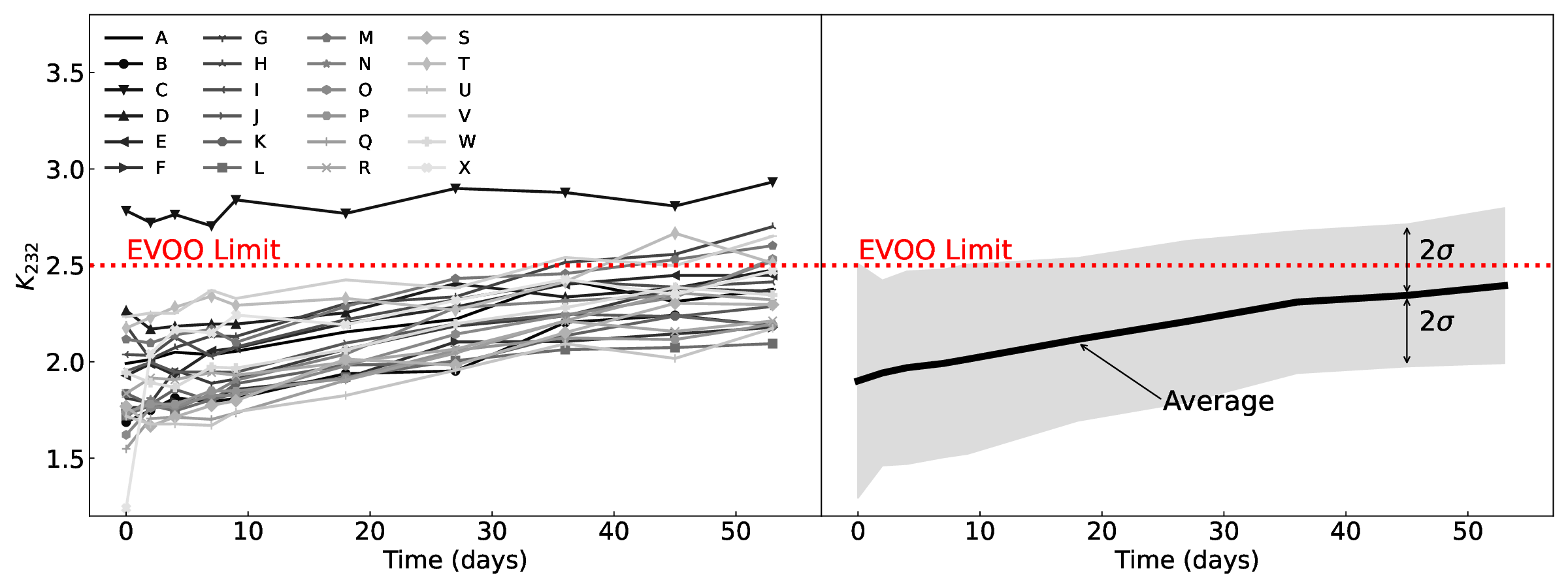}
    \caption{UV-Spectroscopy $K_{232}$ parameter over the time during the ageing at 60 $^{\circ}$C. The dark black line marks the average of all the oils. The red line is the maximum value according to the European Commission norm.}
    \label{fig:K232}
\end{figure}

\begin{figure}[hbt]
    \centering
    \includegraphics[width=\textwidth]{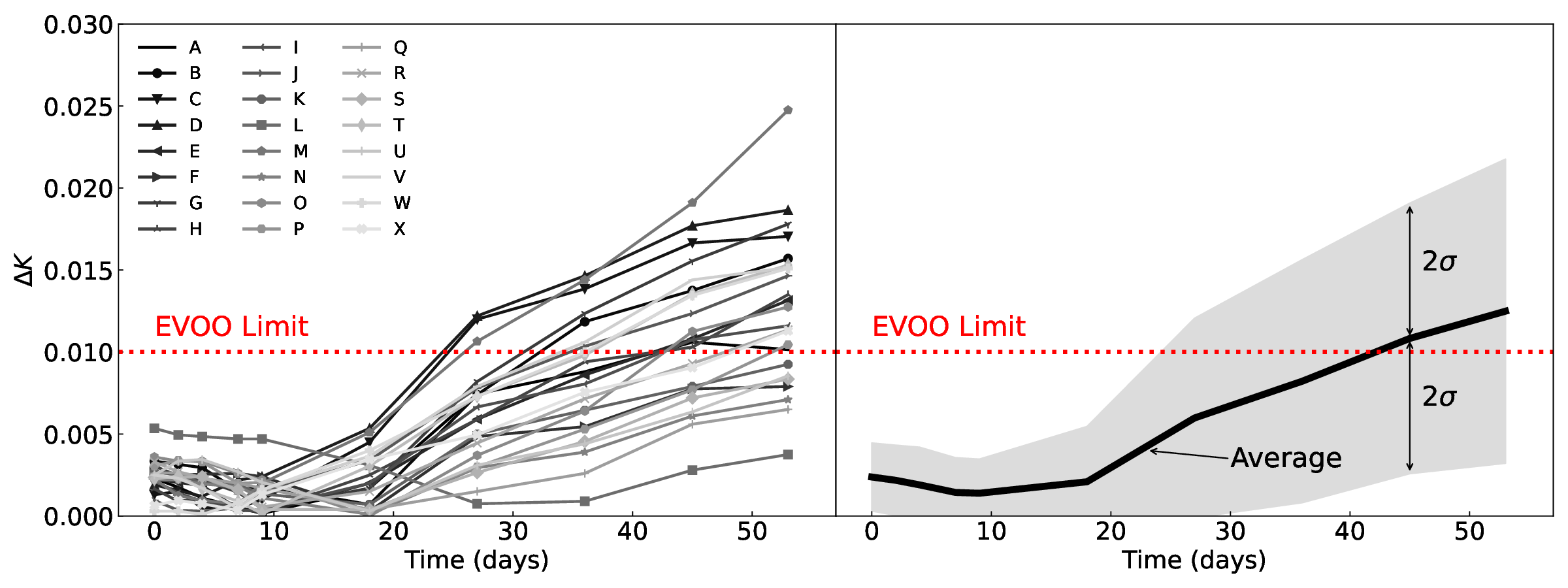}
    \caption{UV-Spectroscopy $\Delta K$ parameter over the time during the ageing at 60 $^{\circ}$C. The dark black line marks the average of all the oils. The red line is the maximum value according to the European Commission norm.}
    \label{fig:DeltaK}
\end{figure}

The evolution of the UV extinction coefficients is shown in Figures \ref{fig:K268},  \ref{fig:K232}, and \ref{fig:DeltaK}.
The first observation is that, generally, the rates of increase of the three parameters vary significantly between the samples. This different ageing patterns may depend, among other factors, on the cultivar of the olive oils used, on the processing method, and the geographical origin of the fruits. It is to be noticed that the EVOO labeled C was characterized by a strong absorption in the UV already before the start of the ageing in this study, that is in fresh conditions. This indicates that the specific bottle had already suffered significant ageing and was no longer an EVOO at the beginning of this study. This can be easily seen in Figure \ref{fig:K232}.

Looking at the single extinction coefficients, Fig. \ref{fig:K268} shows the values of $K_{268}$ for all samples and the average (as a thick solid black line in both panels). Although the average value is not chemically significant per se, it helps determine if there is a common behaviour despite the heterogeneous nature of the EVOOs. The results indicate that no significant changes are visible in the first 10 days, whereas a clear increase is visible after 10 days within the measurement errors. Only 4 oils have a $K_{268}$ parameter that is below the threshold after 53 days (indicating still a high quality). The average rate of change is 0.004/day, the total change of the average over 53 days is 143 \%.
Similarly, Fig. \ref{fig:K232} shows that the values of $K_{232}$ also tend to increase over time, although the change is less pronounced. At the end of the study, only seven oils have exceeded the regulatory limit for EVOO. The average rate of change is 0.008/day and the change of the average over 53 days is 26\%.

The behaviour of $\Delta K$ (Fig. \ref{fig:DeltaK}) mimics the one of $K_{268}$, showing an appreciable increase only after 10-15 days. The uncertainty on the value of $\Delta K$ was estimated by repeating the measurement with the same oil at the same ageing step and was fount to be approximately 0.02 ($2 \sigma$). Since the $\Delta K$ values are comparable to the measurement error, it was decided not to use them as labels for the machine learning model.

The increase in $K_{268}$ and $K_{232}$ over time is consistent with the observations previously reported \cite{mancebo2008kinetic,conte2020temperature,escudero2016influence}. It is important to note that, differently from previous studies, the trends presented here arise from a large variety of samples and are therefore not specific of a cultivar or of a geographical origin. Additionally, unlike previous studies, the ageing in this work was performed using closed vials with the minimum amount of oxygen possible to mimic the storage conditions of commercial bottles. This precaution avoids the demonstrated acceleration of oxidation and reduction of shelf life associated with oxygen in the headspace \cite{iqdiam2020influence}.

The results of this work indicate that the absorbance at 268 nm ($K_{268}$) is the most sensitive parameter between the three to detect ageing due to temperature. This is consistent with the fact that the $K_{268}$ extinction coefficient is an indicator of the presence of secondary oxidation products \cite{escudero2016influence}.


\subsection{Fluorescence Spectroscopy Results}
\label{sec:results_FL_spectra}

\begin{figure}[b!]
    \centering
    \includegraphics[width=\textwidth]{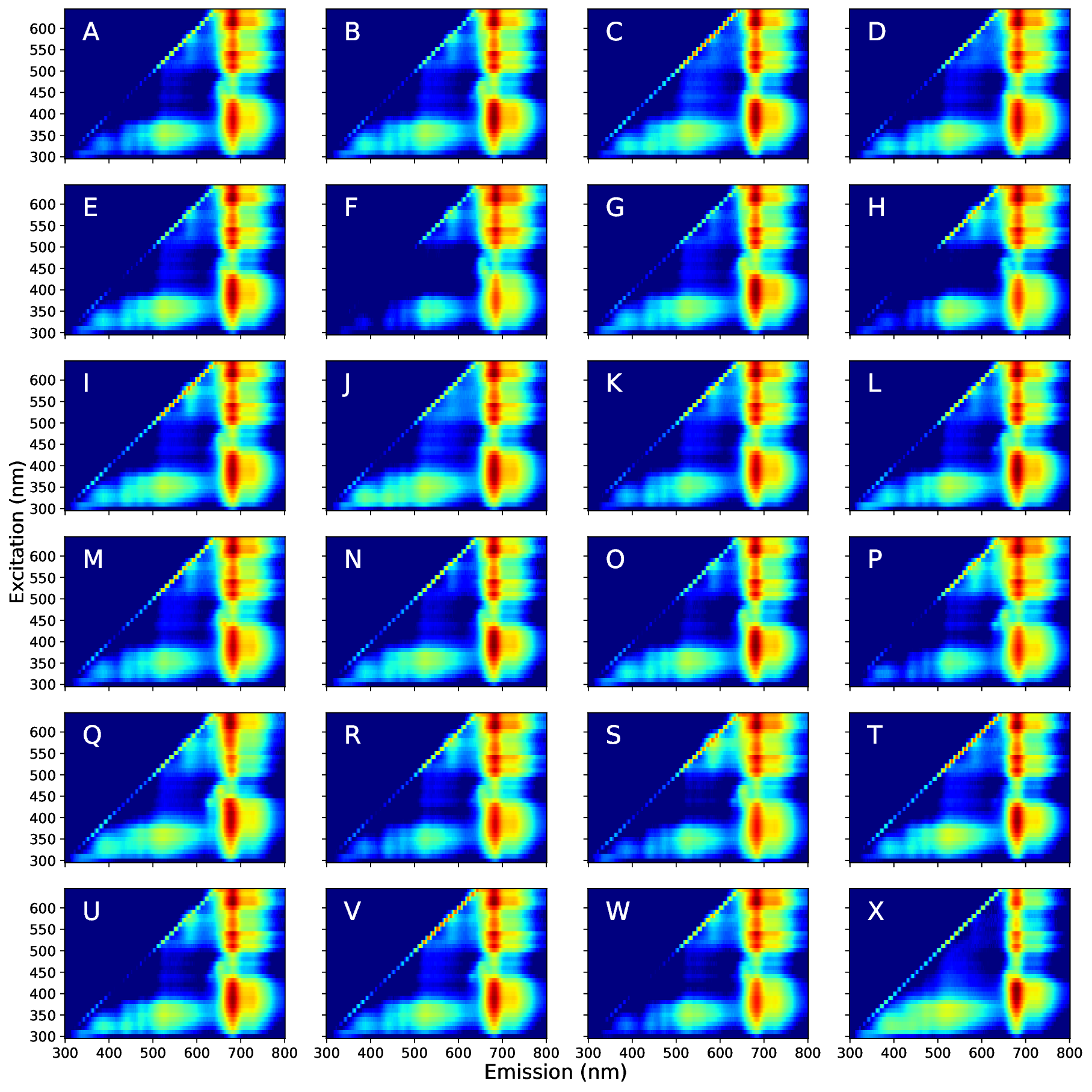}
    \caption{Excitation-emission matrices (EEMs) of the 24 EVOO at ageing step 0 (fresh oils). The data are shown in logarithmic scale to make differences more visible. Figure from \cite{venturini2023}.}
    \label{fig:EEM}
\end{figure}

The EEMs of the oils measured immediately after opening the bottles (Fig. \ref{fig:EEM}) have strong similarities with a small variability in the fluorescence characteristics of specific spectral regions. The strongest signal is clearly the asymmetric band with emission between ca. 650 nm and 750 nm. This band is strongest for excitation wavelengths of 350-400 nm and 600-650 nm and is attributed to chlorophyll pigments, consistent with what has previously been reported in the literature \cite{al2021cultivar,sikorska2012analysis,zandomeneghi2005fluorescence,baltazar2020development}.
The most evident differences between the oils arise for excitations between 310 and 350 nm end emissions between 350 nm and 600 nm, such as pyridoxine, vitamins and flavins \cite{zandomeneghi2005fluorescence}. It is difficult to assign these bands because there are many endogenous molecules which fluoresce in this spectral range. Additionally, oxidation products that may be present in fresh olive oil also emit in a spectral range of 400 to 450 nm \cite{baltazar2020development,al2019preliminary}. 
It is to be noticed that the measurements were performed on undiluted samples. The reason is that in this work the authors want to identify a method which is as simple as possible, without any sample handling, to be usable in field conditions and without special laboratory equipment and knowledge. Therefore, even if the spectra may be subjected to the inner filter effect \cite{skoog2017principles}, this is either not relevant or can be exploited to be more sensitive to thermal degradation.

\begin{figure}[t!]
    \centering
    \includegraphics[width=\textwidth]{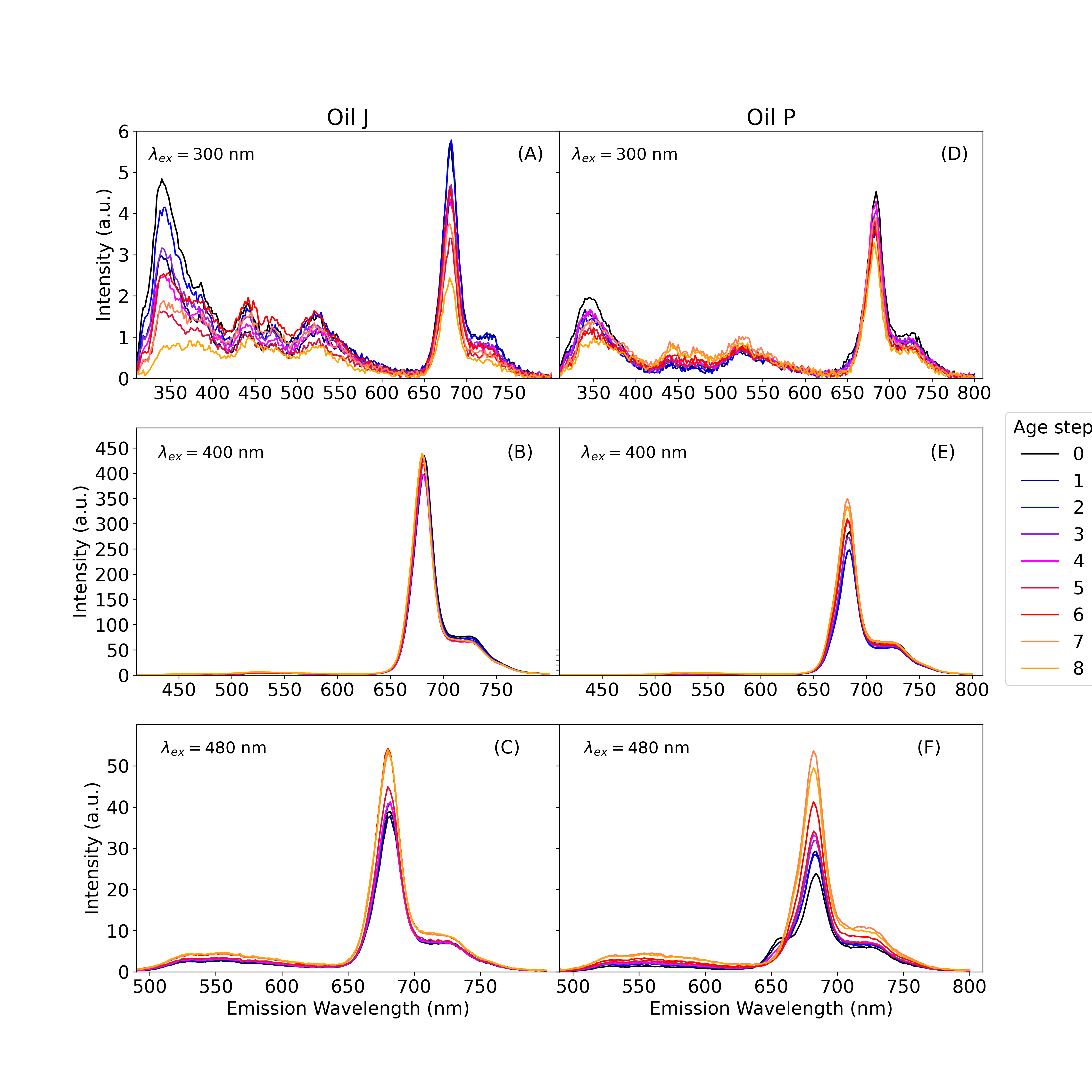}
    \caption{Evolution of the fluorescence emission with the ageing for two selected oils. Left panels: oil J; right panels: oil P. $\lambda_{ex}$ is the excitation wavelength.}
    \label{fig:spectra_evol}
\end{figure}

The evolution of the fluorescence emission spectrum with thermal ageing is shown in Fig. \ref{fig:spectra_evol} for two selected oils and at three excitation wavelengths. The excitation wavelengths 300 nm and 480 nm were chosen because they correspond to those selected by the information content maximization algorithm described in Section \ref{sec:wavelength} (see the discussion in Section \ref{ref:sec:feat_a}). 400 nm was chosen because it corresponds to the absorption band of chlorophylls, which have the strongest fluorescence signal, as it can be seen from Fig. \ref{fig:EEM}.

The analysis of the changes shows that there are some common changes and some oil-specific ones. For an excitation wavelength $\lambda_{ex}=300$ nm, there is a decrease in the intensity of both peaks at ca. 350 and at ca. 680 nm for all oils. This can be interpreted as an oxidative reduction of tocopherols and chlorophylls, respectively. On the other hand, the changes in the region between 400 and 500-550 nm depend on the oil chemical properties, showing an increase (for example for oil F, O and P, the latter shown in Figure \ref{fig:spectra_evol} in panel (D)), no change (for example for oil A or H, not shown here), or an unclear evolution (for example for oil J shown in panel (A) in Figure \ref{fig:spectra_evol}). This specificity can be interpreted as the result of the formation of primary and secondary oxidation products which are expected to be very different in the various samples due to the heterogeneity of the oil cultivar and geographical origin.
For the excitation wavelength $\lambda_{ex}=400$ nm there is a general tendency in the increase of the peak at ca. 680 nm (panel (E) (oil P in panel (E) in Figure \ref{fig:spectra_evol}), but such a behaviour is less pronounced in some oils (for example, oil J, panel (B) in Figure \ref{fig:spectra_evol}).
Significant changes in the fluorescence spectra were observed for the excitation wavelength $\lambda_{ex}=480$ nm (panels (C) and (F) in Figure \ref{fig:spectra_evol}). Here there is a consistent increase in the intensity, which is observed for all the oils investigated both in the region between 500 nm 625 nm and of the peak at ca. 680 nm. In some of the oils the shape of the spectra also show a significant change (for example oil P shown in panel (F) in \ref{fig:spectra_evol}). These results can be interpreted as the result of the formation of oxidation products, the difference in the absorption and emission characteristics of chlorophyll $a$ and $b$ and pheophytin $a$ and $b$ \cite{galeano2003simultaneous} which may be affected differently by the thermal degradation, in combination with the inner filter effect \cite{torreblanca2019laser}.

\begin{figure}[t!]
    \centering
    \includegraphics[width=10 cm]{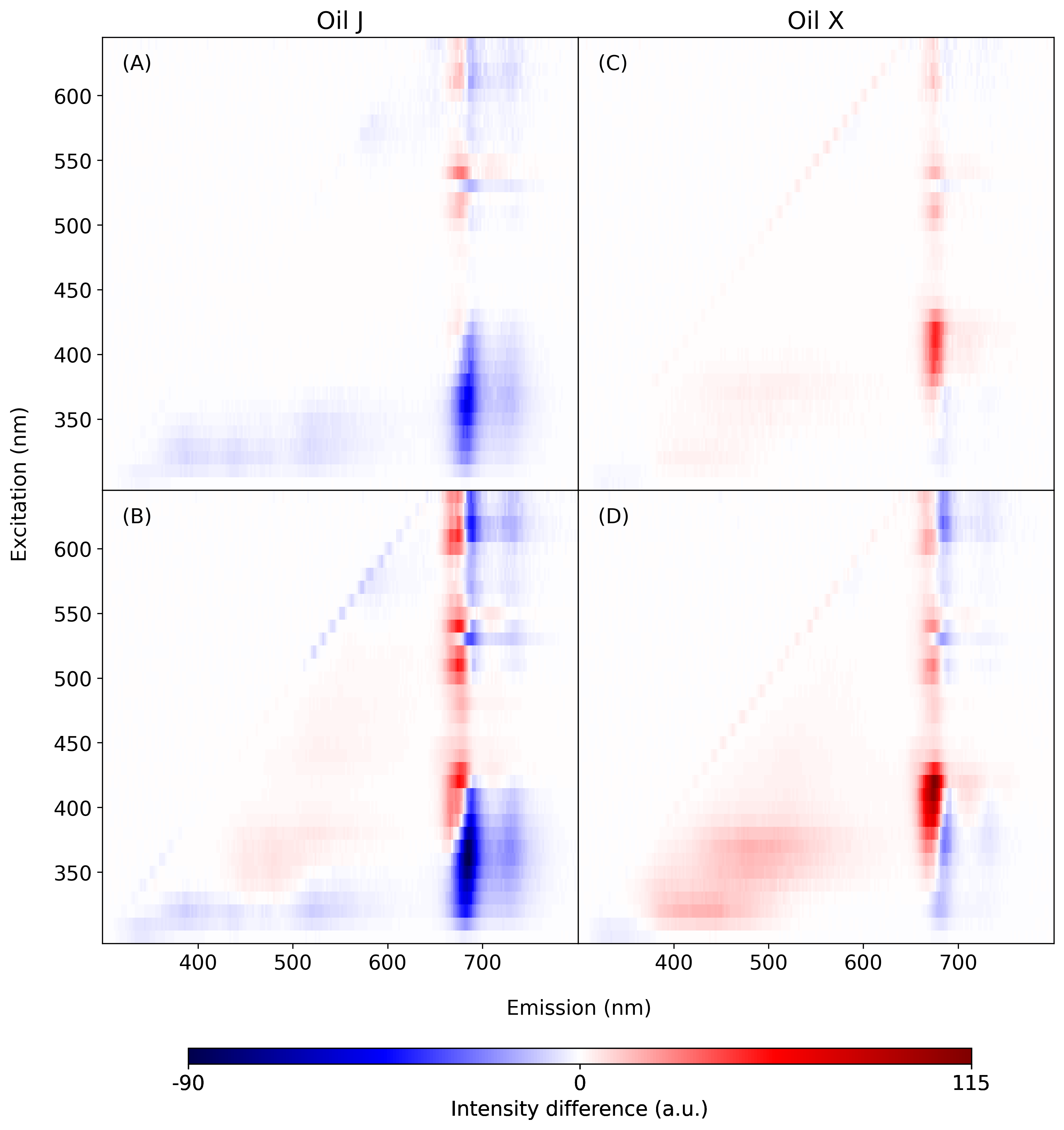}
    \caption{Difference of the florescence intensity at selected ageing step and fresh for two selected oils. Red indicates positive values, blue negative ones. Left panels: oil J; right panels: oil X. Panels A, C: Difference of the intensity after 9 days and fresh; Panels B, D: Difference of the intensity after 53 days and fresh. The intensity is plotted in logarithmic scale to make differences more visible.}
    \label{fig:agein_evol}
\end{figure}

The spectral regions where the fluorescence undergoes the strongest change can be visualized by calculating the difference of the fluorescence intensity at a given ageing step and in fresh condition. This is shown in Fig. \ref{fig:agein_evol} for two selected oils at two ageing steps, namely after 9 and 53 days. In both oils there is an increase (red) of the fluorescence intensity for excitations above 400 nm and emissions between 650 and 680 nm, whereas the intensity of the shoulder between 680 nm and 700 nm tends to decrease, particularly for excitations below 400 nm. Other smaller changes are more specific to the oil and reflect the different chemical composition and heterogeneity of the oils investigated.

\subsection{Feature analysis} 
\label{ref:sec:feat_a}

 \begin{figure}[t!]
    \centering
    \includegraphics[width=\textwidth]{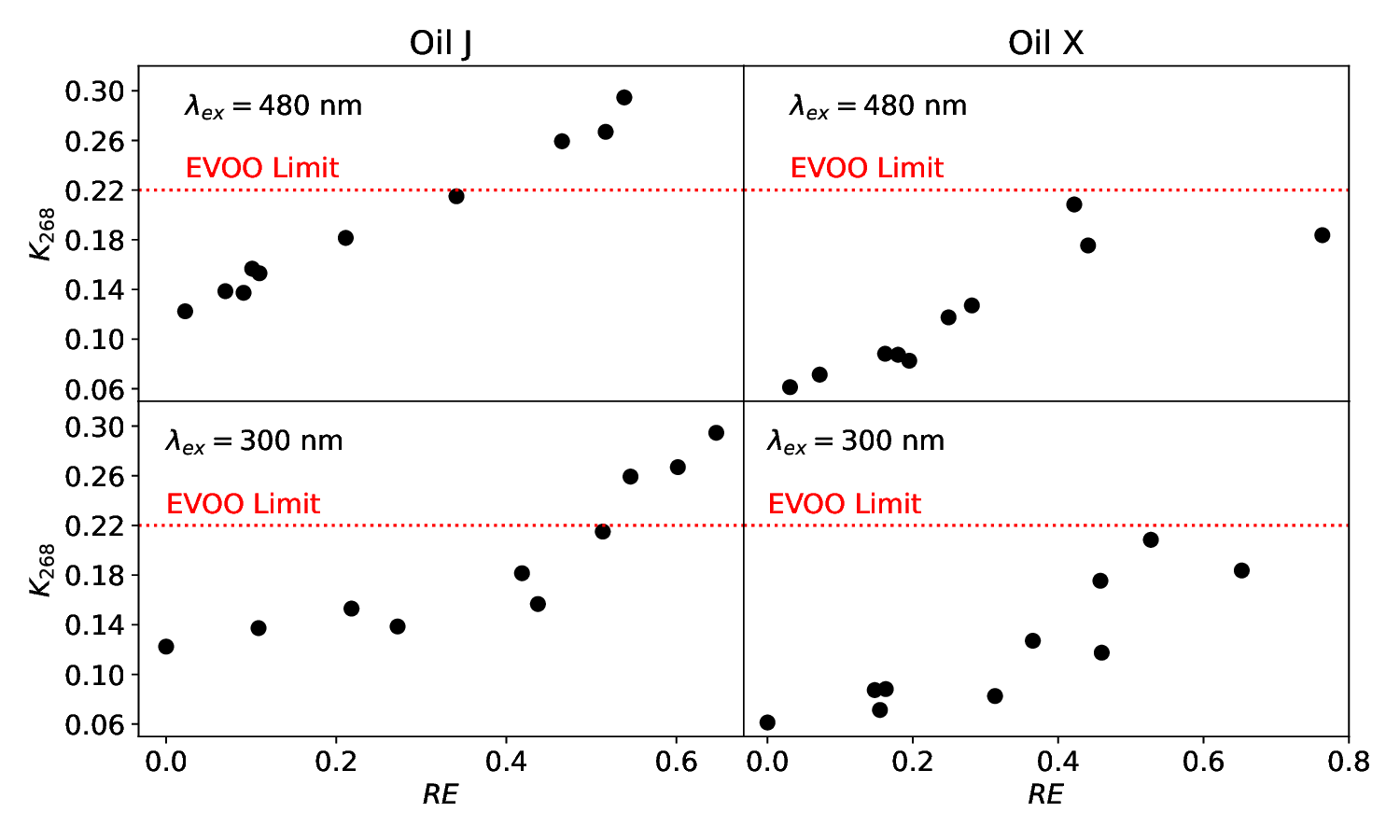}
    \caption{Evolution of the UV parameter $K_{268}$ versus the relative error with at the two wavelengths containing the most information. Left panels: oil X; right panels: oil j. Top panels: $\lambda^{[2]}=$ 300 nm; $\lambda^{[1]}=$ 480 nm. The limits allowed for EVOO are also shown in red.}
    \label{fig:K268vsRE}
\end{figure}

Using the relative error $RE$ calculated according to Eq. \ref{eqn:RE}, the most relevant wavelengths resulted $\lambda^{[1]}= 480$ nm and $\lambda^{[2]} = 300$ nm, which appeared 15 and 6 times respectively. The remaining three oils showed maximal $RE$ at 320 nm (twice) and 450 nm (once). It is to be noted that for oils in which $RE$ did not result highest at 480 nm, for example $\lambda^{[1]}_{C} = 450$ nm,  the second most relevant wavelength in terms of $RE$ was always 480 nm. This motivated the selection $\lambda^{[1]}= 480$ nm and $\lambda^{[2]} = 300$ nm for all the oils. The results can be interpreted with the help of Fig. \ref{fig:spectra_evol}: for both $\lambda^{[1]}= 480$ nm and $\lambda^{[2]} = 300$ the spectra undergoes significant changes. Since the $RE$ is normalized by the norm squared, the relative changes have a more important role than the absolute intensity, which is much stronger for excitation of 400 nm (corresponding to the absorption of chlorophylls), as discussed in Section \ref{sec:results_FL_spectra}.

The $RE$ allows to identify which excitation wavelength has the overall strongest impact upon ageing on the fluorescence spectrum. To be usable for the classification of oil quality, however, $RE$ must correlate with changes in the UV parameters. Figure~\ref{fig:K268vsRE} shows the coefficient $K_{268}$ vs. relative errors $RE(480)$ and $RE(300)$ for two selected oils, and demonstrates that this is indeed the case. The evolution of the UV parameter $K_{232}$ shows a similar behaviour (see Additional Material).

\subsection{Predictive models} 

\label{sec:res:classifier}
\begin{figure}[hbt]
\centering
\subfloat[Results for the classification done with the class derived from the parameter $K_{268}$.]{%
  \includegraphics[clip,width=0.8\columnwidth]{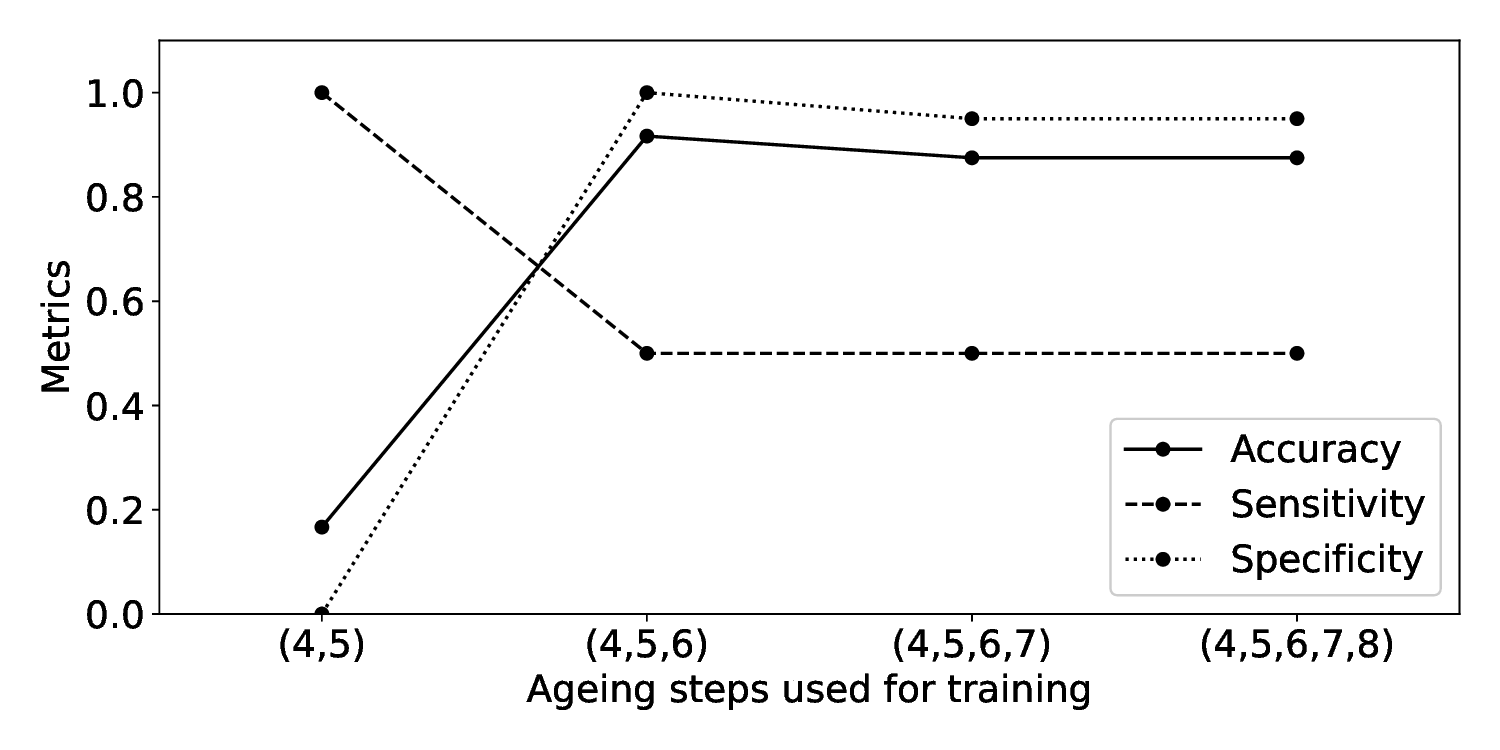}%
}

\subfloat[Results for the classification done with the class derived from the parameter $K_{232}$.]{%
  \includegraphics[clip,width=0.8\columnwidth]{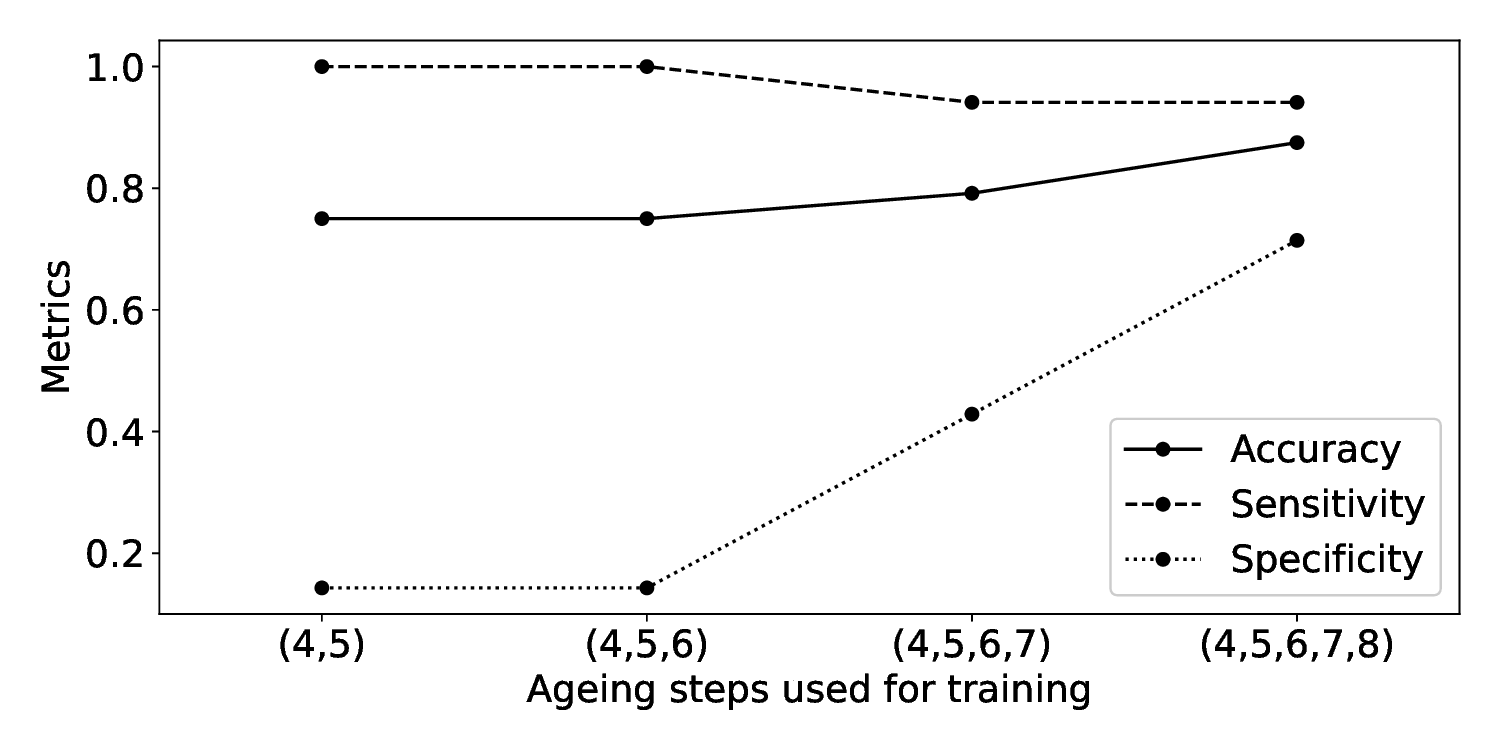}%
}

\caption{Results obtained for the prediction of the classes at ageing step 9 (Method 1).}
\label{fig:val_step9}
\end{figure}

The performance of the model is quantified in terms of accuracy, sensitivity, and specificity \cite{hastie2009multi}. The results are shown in Figure \ref{fig:val_step9}, where the metrics are plotted for all the different training listed in Table \ref{tab:models} and for the classes obtained from the parameters $K_{268}$ in panel (a) and $K_{232}$ in panel (b), respectively. The results show that the model is able to detect oils that have passed the thresholds with a high accuracy, above 90$\%$. 
As expected, the greater the size of the data set (or, in other words, the more aging steps employed during training), the better the performance becomes. When using the parameter $K_{268}$ to label the class (see panel (a) in Figure \ref{fig:val_step9}), already with three ageing steps (in particular 4,5 and 6) the performance is exceptional, with an accuracy of 90$\%$. The results are consistent with the fact that the $RE$ grows noticeably only after the first ageing steps, thus including step 6 give the model enough information on the evolution of the ageing process. This is not reflected when predicting the class from the parameters $K_{232}$. In this case the best results are obtained when using all available ageing steps (4,5,6,7 and 8) as can be seen from panel (b) in Figure \ref{fig:val_step9}. This can be understood in term of the less sensitivity of this parameter to thermal oxidation as discussed previously.

\begin{figure}[hbt]
\centering
\subfloat[Results for the classification done with the class derived from the parameter $K_{268}$.]{%
  \includegraphics[clip,width=0.8\columnwidth]{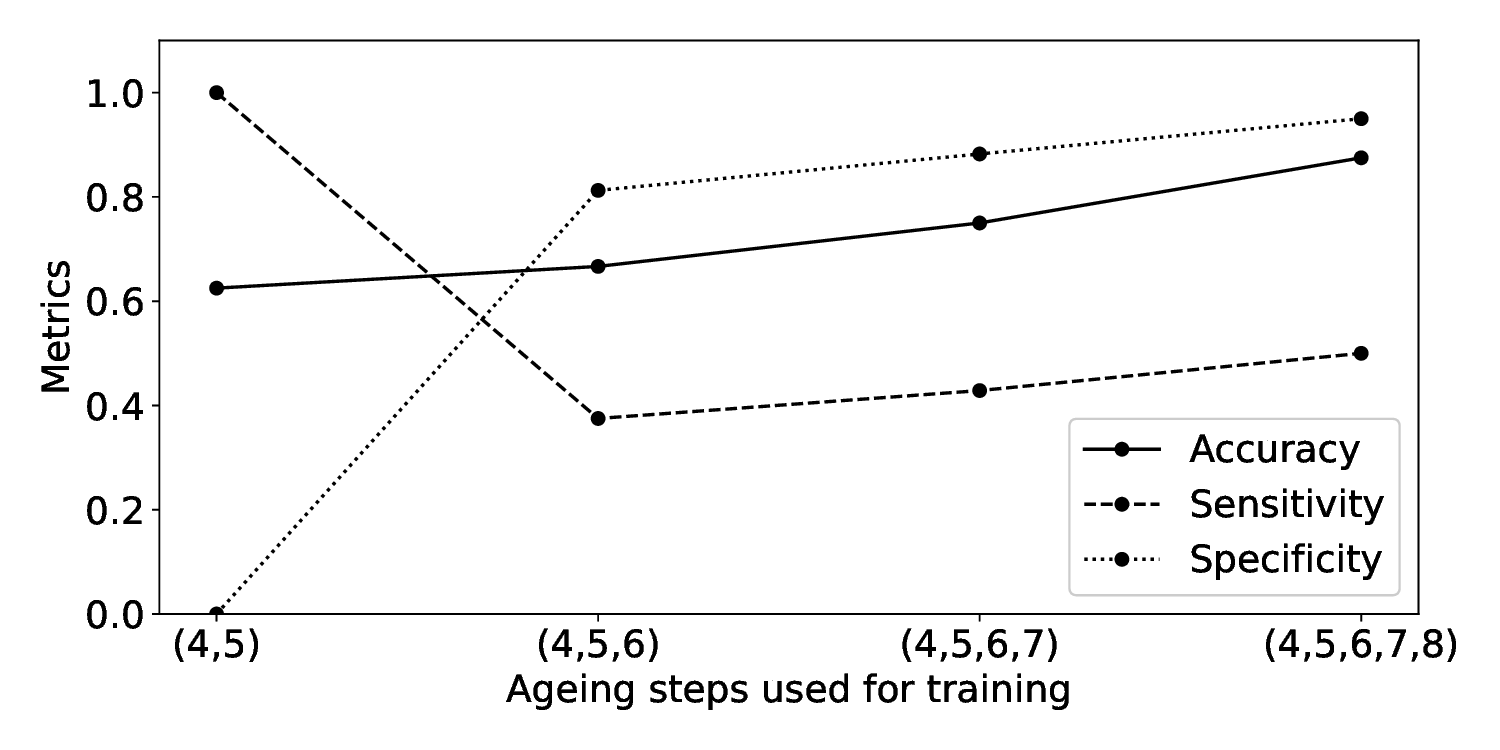}%
}

\subfloat[Results for the classification done with the class derived from the parameter $K_{232}$.]{%
  \includegraphics[clip,width=0.8\columnwidth]{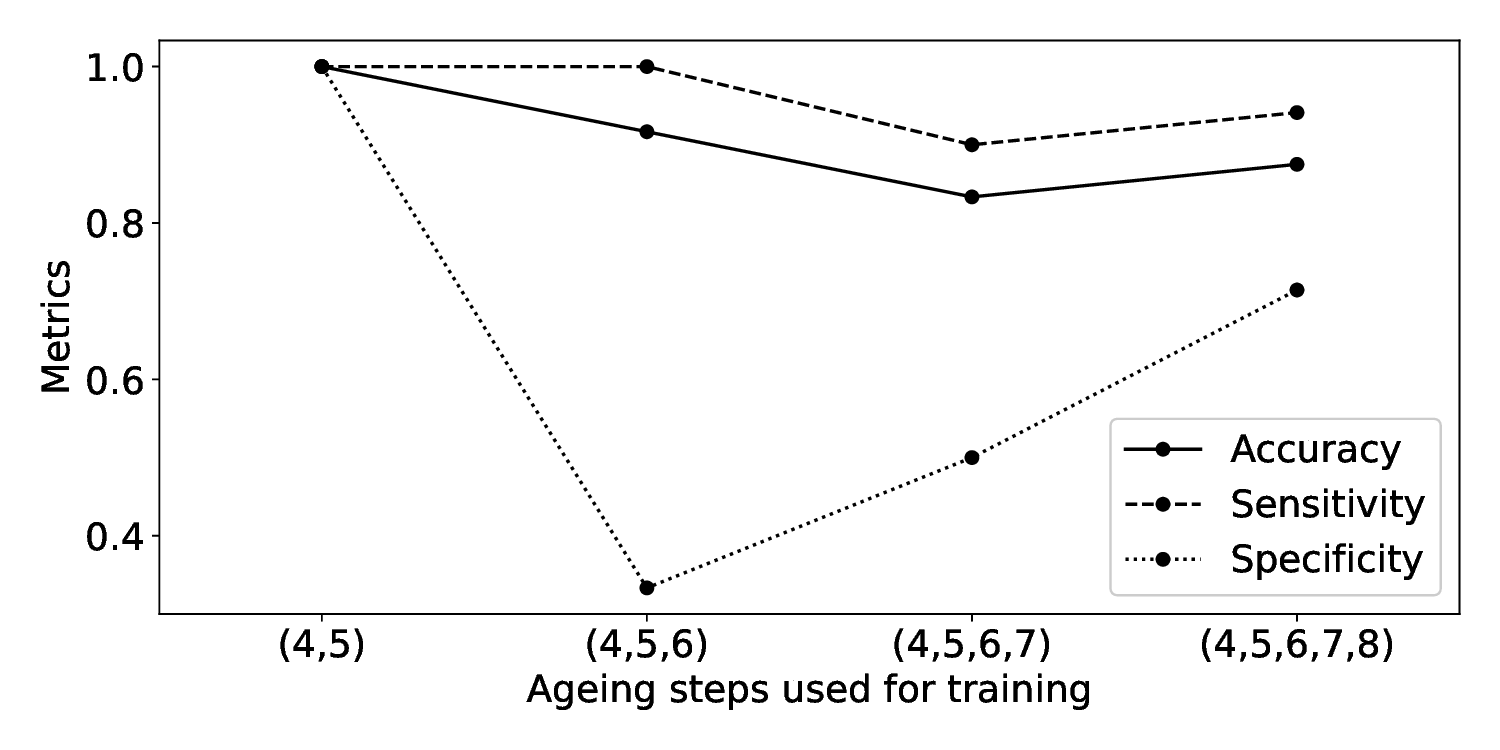}%
}

\caption{Results obtained for the prediction of the immediate next ageing step (Method 2).}
\label{fig:val_step_next}
\end{figure}

The values of the metrics are summarized in Table \ref{tab:results} for clarity. The values in the table correspond to the training performed using the largest number of ageing steps. It should be noted that the metric values are the same in both cases. The reason is that the dataset has only 24 oils, thus the accuracy, for example, can only assume a discrete set of values (0, 1/24, 1/12, 3/24, etc.).

\begin{table}[h]
  \centering
  \begin{tabular}{c|c|c}
  \specialrule{.2em}{0em}{0em} 
  \multicolumn{3}{c}{Predictions on ageing step 9} \\ 
  \specialrule{.2em}{0em}{0em} 
   Metric &
   \specialcell{\textbf{Classes derived}\\\textbf{from $K_{268}$} } & \specialcell{\textbf{Classes derived}\\\textbf{from $K_{232}$} } \\ 
    \specialrule{.1em}{0em}{0em}
    Accuracy & 0.92 & 0.88 \\
    Sensitivity & 0.5 & 0.94 \\
    Specificity & 0.95 & 0.71 \\
  \specialrule{.2em}{0em}{0em}
  \multicolumn{3}{c}{Predictions on next ageing step} \\ 
  \specialrule{.2em}{0em}{0em} 
    Accuracy & 0.92 & 0.88 \\
    Sensitivity & 0.5 & 0.94 \\
    Specificity & 0.95 & 0.71 \\
  \specialrule{.1em}{0em}{0em}
  \end{tabular}
  \caption{Metrics for the AdaBoost model described in the text. Note that the values are the same, since the number of oils that are over or under the threshold is very small and thus the metrics can assume a limited number of values.\label{tab:results}}
\end{table}


The results where the validation is done only on the next ageing step (\textsl{method 2} in Table \ref{tab:models}) are shown in Fig. \ref{fig:val_step_next} for the class obtained from $K_{268}$ and $K_{232}$ in panels (a) and (b) respectively.

From figure \ref{fig:val_step_next}, and from the comparison with Figure \ref{fig:val_step9}, it can be concluded that having enough ageing steps is a pre-requisite for a good prediction. The best results are obtained when using ageing steps 4,5,6,7 and 8 as input for the training. 

The first point in Fig. \ref{fig:val_step_next} in panel (b) requires a discussion. All three metrics (accuracy, sensitivity, and specificity) are equal to 1. This would indicate a perfect prediction, but it is in reality to be attributed to a random effect due (most probably) to the specific data distribution. If this would be a real effect, this perfect prediction would also be visible when additional ageing steps are added to the training data. This is not clearly the case, as can be seen in panel (b) in Fig. \ref{fig:val_step_next}. Thus, this point should not be considered as indicating that a model can predict the quality of oil perfectly.

\section{Conclusions}
\label{sec:conclusions}

In conclusion, this work makes several significant contributions to the study of the quality assessment of extra virgin olive oil (EVOO). Four key contributions can be identified. 

Firstly, the work involves extensive UV-absorption spectroscopy analysis at the specific wavelengths required for the quality assessment specified by the European Regulations over time for a range of commercially available EVOOs. The measurements of three parameters ($K_{232}$, $K_{268}$, and $\Delta K$) provide insights into the oxidation processes 
 giving rise to absorption in the UV range, which is crucial for assessing EVOO's quality.

Secondly, the study conducts extensive total fluorescence spectroscopy analysis of the same oil samples through the acquisition of excitation-emission matrices (EMMs). By measuring the fluorescence intensity at different excitation and emission wavelengths, the study explores the fluorescence characteristics of EVOOs during different ageing stages.

Thirdly, the research identifies the two excitation wavelengths (480 nm and 300 nm) that exhibit the maximum relative change in fluorescence for the majority of the oils. This finding highlights specific wavelengths that are highly informative in terms of assessing EVOO quality based on fluorescence properties.

Lastly, the work proposes a method for the prediction of olive oil quality based on aggregated data. By considering the fluorescence intensity at the two identified excitation wavelengths, the study employs machine learning algorithms (such as AdaBoost, Random Forest, Logistic Regression, and Naïve Bayes) to classify EVOOs as either extra virgin or non-extra virgin based on the UV-spectroscopy criteria defined by the European regulations with an accuracy above 90$\%$. The advantage of the proposed method using aggregated data, is that it does not require a spectrometer or spectral analysis for the classification of EVOO: the realization of a portable device would only require two LEDs for the excitation at the two identified wavelengths, and one photodiode for the collection of the integrated fluorescence spectrum. Clearly, the aggregation process does not allow an in depth analysis of the detailed changes in the oil due to the oxidation process. This could be the subject of a further study. Also, being the oils used in this study very heterogeneous in fruits used for the extraction, geographical origin and advertised quality, the authors expect the results to be quite general in validity, thus with a broad range of applicability. 

Overall, the contributions of this work advance our understanding of EVOO quality assessment by combining UV-absorption spectroscopy, fluorescence analysis, and machine learning techniques. The findings provide valuable insights for the development of low-cost and portable methods for assessing the quality of EVOO, which can have significant implications for the olive oil industry and consumers.

\section{Funding}
This research was supported by the Hasler Foundation project ``ARES: AI for fluoREscence Spectroscopy in oil''.



\bibliographystyle{elsarticle-num-names}
\bibliography{mybibfile}

\end{document}